\documentclass{article}

\usepackage{arxiv}

\usepackage[utf8]{inputenc} 
\usepackage[T1]{fontenc}    
\usepackage{hyperref}       
\usepackage{url}            
\usepackage{booktabs}       
\usepackage{amsfonts}       
\usepackage{nicefrac}       
\usepackage{microtype}      
\usepackage{lipsum}		
\usepackage{graphicx}
\usepackage{doi}
\usepackage{amsmath}
\usepackage{placeins}
\usepackage{tikz}
\usetikzlibrary{positioning, decorations.pathreplacing, calligraphy, calc}

\author{Bouarfa Mahi \\ Quantiota \\ Email: info@quantiota.org}

\title{Structured Knowledge Accumulation: An Autonomous Framework for Layer-Wise Entropy Reduction in Neural Learning}

\author{ \href{https://orcid.org/0009-0008-7158-2729}{\includegraphics[scale=0.06]{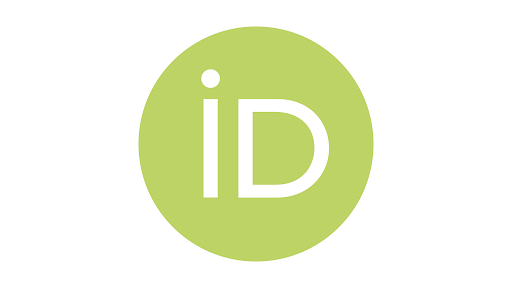}}\hspace{1mm}Bouarfa Mahi Quantiota \thanks{Use footnote for providing further
		information about author (webpage, alternative
		address)---\emph{not} for acknowledging funding agencies.} \\
	Université Joseph Fourier\\
Grenoble, Auvergne-Rhône-Alpes, FR\\
	\texttt{info@quantiota.org} }

\renewcommand{\shorttitle}{An Autonomous Framework for Layer-Wise Entropy Reduction}

\hypersetup{
pdftitle={A template for the arxiv style},
pdfsubject={q-bio.NC, q-bio.QM},
pdfauthor={David S.~Hippocampus, Elias D.~Striatum},
pdfkeywords={First keyword, Second keyword, More},
}

\begin{document}
\maketitle

\begin{abstract}
	We introduce the Structured Knowledge Accumulation (SKA) framework, which reinterprets entropy as a dynamic, layer-wise measure of knowledge alignment in neural networks. Instead of relying on traditional gradient-based optimization, SKA defines entropy in terms of knowledge vectors and their influence on decision probabilities across multiple layers. This formulation naturally leads to the emergence of activation functions such as the sigmoid as a consequence of entropy minimization. Unlike conventional backpropagation, SKA allows each layer to optimize independently by aligning its knowledge representation with changes in decision probabilities. As a result, total network entropy decreases in a hierarchical manner, allowing knowledge structures to evolve progressively. This approach provides a scalable, biologically plausible alternative to gradient-based learning, bridging information theory and artificial intelligence while offering promising applications in resource-constrained and parallel computing environments.
\end{abstract}


\section{Introduction}
The role of entropy in intelligent systems is crucial in how one interprets the underlying structure and functioning of AI. Deep learning, especially the deep superposition of gradients, has been aided by traditional learning paradigms that have proven extremely effective, such as the ones based on gradient backpropagation. Unfortunately, these methods are computationally expensive, biologically unrealistic, and impenetrably opaque. There needs to be an attempt to overcome the divide between the theoretical tenets of information theory and its actual engineering implementation in neural networks. In view of this problem, we present a novel framework that treats the redefinition of entropy as a continuous process of structured knowledge accumulation, which enables a different approach to the internal organization of AI learning system.

Entropy, as classically defined by \cite{shannon1949mathematical}, is given as:
\begin{equation}\label{eq1}
H = - \sum p_i \log_2 p_i,
\end{equation}
which measures uncertainty in a discrete, static probabilistic system. While this framework serves as the basis of information theory, it falls short of getting the continuous and dynamic structuring of intelligent systems knowledge such as neural networks. The sigmoid function which is defined as:
\begin{equation}
\sigma(z) = \frac{1}{1+e^{-z}},
\end{equation}
has been a keystone of AI. This function is commonly used for activation in neural networks. However, its theoretical justification beyond empirical utility is  an open question.

Conventional training through backpropagation spreads errors backward via the network, which requires additional computational means while lacking limiting scalability, biological plausibility, and real-world utilization. Tackling these limitations, this article establishes the SKA framework, which redefines entropy as a hierarchical and continuous process of knowledge accumulation.

We propose:
\begin{enumerate}
    \item Entropy as a dynamic measure, expressed in layers as:
    \begin{equation}\label{eq3}
    H^{(l)} = -\frac{1}{\ln 2} \sum_{k=1}^{K} \mathbf{z}^{(l)}_k \cdot \Delta \mathbf{D}^{(l)}_k,
    \end{equation}
    approximating the continuous entropy formulation:
    \begin{equation}\label{eq4}
    H = -\frac{1}{\ln 2} \int z \, dD.
    \end{equation}
    
    \item Cumulative knowledge \(\mathbf{z}\) serves as the building block for literacy.

\item Learning rules involving minimization of local entropy and lack of backpropagation.
\end{enumerate}

We formulate the sigmoid function as an emergent phenomenon of continuous entropy flow and transfer this paradigm to fully connected networks, showing how learning emerges from the micromatching of knowledge with the decision-making dynamics. This framework offers a biologically accurate, scalable explanation of deep learning that displacement methods, increases the optimization effectiveness, and adds optimization transparency. 

The creation of AI systems, neuroscience, and computational intelligence would all benefit from understanding knowledge accumulation through entropy dynamics. With this information, SKA is able to streamline neural network training in resource-limited environments, promote concurrent learning, and aid in the formulation of more interpretable AI systems. The framework further motivates the development of energy-efficient and biologically inspired computing systems, making it a fundamental step toward the next generation of intelligent architectures.

\section{Literature Review}
The study of AI and neural networks has greatly benefitted from the research done on knowledge-based entropy systems. Early works of \cite{Shannon1948} proposed the concept of entropy in information theory, which has since been a driving force for many subsequent learning paradigms in neural networks. Modern research developed an entropy-based framework to improve neural networks and create knowledge alignment. The assessment of entropy has been studied in several different ways. \cite{Shalev2022} studied a method of neural joint entropy approximation which contributes to how entropy can be leveraged for learning efficacy. \cite{Oizumi2015} analyzed integrated information from the perspective of decoding, which proposes a way to assess the level of complexity present in neural representations. Additionally, some of this research concentrates on learning theories with biological motivation. \cite{Weng2022} proposed notion networks that are inspired by the brain, demonstrating learning from watching scenes unfold in a disorganized fashion, emulating human perception. \cite{Leung2021} examined the collapse of integrated information mechanisms due to anesthetic loss which sheds light on the processing of neural information.\\
\indent Multiple difficulties stem from using backpropagation-based learning, including the lack of biological plausibility as well as computational inadequacy. These concerns have driven the exploration of alternative optimization methods. \cite{Xie2017} proposed Generative ConvNets, which are energy-based models capable of synthesizing dynamic patterns and investigates into local learning mechanisms. Similarly,  \cite{Xie2003} illustrate the relationship between contrastive and backpropagation Hebbian learning, which offer a more believable biological framework. \cite{Koulakov2023}, regarding the encoding of innate ability via a genomic bottleneck, investigated how biological concepts can be integrated into machine learning systems. \cite{Lagergren2020} built on this idea using biologically-informed neural networks to guide mechanistic modeling based on sparse experimental data. \\
\indent The last several years have witnessed considerable progress in optimizing the training processes for deep neural networks. \cite{Defazio2014} presented SAGA, an incremental gradient optimization technique that enhances convergence and lowers variance. \cite{Lin2015} pioneered acceleration catalysts also known as first-order convex optimization acceleration in the context of machine learning. The vanishing gradients problem remains unresolved in deep neural network architectures, and \cite{He2016} proposed ResNets for addressing this enduring challenge. \cite{Qiu2019} analyzed the use of contrastive divergence for training energy-based latent variable models, which enhances representation learning in neural networks. Adding entropy-based learning yields results beyond those achievable with standard approaches in machine learning. In their study, \cite{shalev2022neural} introduced a neural network-based approach for estimating joint entropy by leveraging mutual information neural estimation techniques. In addition, Hess et al. \cite{hess2023knowledge} studied how continual learning models accumulate knowledge and confront feature forgetting, highlighting that while absolute forgetting may be minimal, relative forgetting can significantly impede the development of robust general representations. In the same context, as discussed by \cite{das2024symbolic}, knowledge distillation techniques have evolved significantly. Furthermore, \cite{xu2019main} introduced a main/subsidiary network framework to simplify binary neural networks by employing a learning-based approach to filter pruning, thereby enhancing efficiency without compromising performance.

\section{Our Obtained Results}

\subsection{Redefining Entropy in the SKA Framework}

Entropy traditionally quantifies uncertainty in probabilistic systems, but its classical form is static and discrete, limiting its applicability to dynamic learning processes like those in neural networks. In the SKA framework, we redefine entropy as a continuous, evolving measure that reflects knowledge alignment over time or processing steps. This section contrasts Shannon’s discrete entropy with our continuous reformulation, enabling the use of continuous decision probabilities and supporting the derivation of the sigmoid function through entropy minimization.

\subsubsection{Classical Shannon Entropy}
For a binary system with decision probability \( D \), Shannon’s entropy is:
\begin{equation}
	H = -D \log_2 D - (1 - D) \log_2 (1 - D).
\end{equation}
Its derivative with respect to \( D \) is:
\begin{equation}
	\frac{dH}{dD} = \log_2 \left( \frac{1 - D}{D} \right).
\end{equation}
This formulation assumes \( D \) is a fixed probability, typically associated with discrete outcomes (e.g., 0 or 1). While foundational, it does not capture the continuous evolution of knowledge in a learning system, where \( D \) may vary smoothly as the network processes inputs. To address this, we seek a continuous entropy measure that accommodates dynamic changes in \( D \), aligning with the SKA’s focus on knowledge accumulation.

\subsubsection{Entropy as a Measure of Knowledge Accumulation}\label{contentropy}
In SKA, we redefine entropy for a single neuron as a continuous process:
\begin{equation}
	H = -\frac{1}{\ln 2} \int z \, dD.
\end{equation}
Here, \( z \) represents the neuron’s structured knowledge, and \( dD \) is an infinitesimal change in the decision probability, treated as a continuous variable over the range \( [0, 1] \). The factor \( -\frac{1}{\ln 2} \) ensures alignment with base-2 logarithms, consistent with Shannon’s information units. Unlike the static snapshot of Equation (1), this integral captures how entropy accumulates as \( z \) drives changes in \( D \), reflecting a dynamic learning process.

For a layer \( l \) with \( n_l \) neurons over \( K \) forward steps, we approximate this continuous form discretely:
\begin{equation}
	H^{(l)} = -\frac{1}{\ln 2} \sum_{k=1}^{K} \mathbf{z}^{(l)}_k \cdot \Delta \mathbf{D}^{(l)}_k, 
\end{equation}
where \( \mathbf{z}^{(l)}_k = [z_1^{(l)}(k), \dots, z_{n_l}^{(l)}(k)]^T \) is the knowledge vector, \( \Delta \mathbf{D}^{(l)}_k = [\Delta D_1^{(l)}(k), \dots, \Delta D_{n_l}^{(l)}(k)]^T \) is the vector of decision probability shifts, and the scalar product is:
\begin{equation}
	\mathbf{z}^{(l)}_k \cdot \Delta \mathbf{D}^{(l)}_k = \sum_{i=1}^{n_l} z_i^{(l)}(k) \Delta D_i^{(l)}(k).
\end{equation}
The total network entropy sums over all layers:
\begin{equation}
	H = \sum_{l=1}^{L} H^{(l)}.
\end{equation}
Equation (4) is the core theoretical construct, with Equation (3) as its practical discrete approximation. As \( K \to \infty \) and \( \Delta \mathbf{D}^{(l)}_k \) becomes infinitesimally small, Equation (3) approaches the continuous integral, enabling us to model \( D \) as a smooth function of \( z \). This continuous perspective is essential for deriving the sigmoid using dynamics in later sections, while the discrete form supports implementation in neural architectures.

\subsubsection{Accumulated Knowledge}
Knowledge accumulates over steps:
\begin{equation}
	z_k = z_0 + \sum_{f=1}^{k} \Delta z_f.
\end{equation}
In a layer, \( \mathbf{z}^{(l)}_k \) evolves, reducing \( H^{(l)} \) as it aligns with \( \Delta \mathbf{D}^{(l)}_k \).
\subsection{Deriving the Sigmoid Function}

The SKA framework posits that the sigmoid function emerges naturally from continuous entropy minimization, linking structured knowledge to decision probabilities. This section demonstrates that when \( D \) follows \( D = \frac{1}{1 + e^{-z}} \), the SKA entropy \( H_{\text{SKA}} \) equals the classical Shannon entropy \( H_{\text{Shannon}} \), differing by a constant (zero). By leveraging the continuous formulation from Section 2, we derive this equivalence, reinforcing the framework’s theoretical grounding.

\subsubsection{Key Definitions}\label{keydef}

\paragraph{Shannon Entropy (for binary decisions):}
For a binary system with continuous decision probability \( D \):
\begin{equation}
	H_{\text{Shannon}} = -D \log_2 D - (1-D) \log_2 (1-D).
\end{equation}

\paragraph{SKA Entropy (layer-wise, for a single neuron):}
The SKA entropy, defined continuously as in Section \ref{contentropy}, is:
\begin{equation}
	H_{\text{SKA}} = -\frac{1}{\ln 2} \int z \, dD,
\end{equation}
where \( z = -\ln\left(\frac{1-D}{D}\right) \) relates knowledge to \( D \), consistent with \( D = \frac{1}{1 + e^{-z}} \) as shown in Section \ref{keydef}.

\subsubsection{Equivalence Proof}

Substituting \( z = -\ln\left(\frac{1-D}{D}\right) \) (or equivalently, \( z = \ln\left(\frac{D}{1-D}\right) \)) into \( H_{\text{SKA}} \):
\begin{equation}
	H_{\text{SKA}} = -\frac{1}{\ln 2} \int \ln\left(\frac{D}{1-D}\right) dD.
\end{equation}
Evaluate the integral with substitution \( u = D \), \( du = dD \):
\begin{equation}
	\int \ln\left(\frac{D}{1-D}\right) dD = D \ln\left(\frac{D}{1-D}\right) + \ln(1-D).
\end{equation}
Substituting back:
\begin{equation}
	H_{\text{SKA}} = -\frac{1}{\ln 2} \left[ D \ln\left(\frac{D}{1-D}\right) + \ln(1-D) \right].
\end{equation}
Rewrite \( \ln\left(\frac{D}{1-D}\right) = \ln D - \ln(1-D) \):
\begin{equation}
	H_{\text{SKA}} = -\frac{1}{\ln 2} \left[ D \ln D - D \ln(1-D) + \ln(1-D) \right].
\end{equation}
Factorize:
\begin{equation}
	H_{\text{SKA}} = -\frac{1}{\ln 2} \left[ D \ln D + (1-D) \ln(1-D) \right].
\end{equation}
Thus:
\begin{equation}
	H_{\text{SKA}} = H_{\text{Shannon}}.
\end{equation}

\subsubsection{Implications}

\begin{itemize}
	\item \textbf{Zero Difference}: The SKA and Shannon entropies are identical (differing by zero) when \( D = \frac{1}{1 + e^{-z}} \), confirming the sigmoid as an emergent property of continuous entropy reduction.
	\item \textbf{Knowledge Alignment}: This equivalence stems from \( z \) structuring \( D \) to minimize uncertainty, as defined in Sections 2 and 3.
\end{itemize}

\subsubsection{Significance}

\begin{enumerate}
	\item \textbf{Theoretical Consistency}: SKA extends Shannon entropy into a continuous, dynamic context while preserving its core properties for sigmoidal outputs.
	\item \textbf{Backpropagation-Free Learning}: Since \( H_{\text{SKA}} = H_{\text{Shannon}} \), layer-wise entropy minimization aligns with classical uncertainty reduction, achieved via forward dynamics alone.
	\item \textbf{Biological Plausibility}: The continuous, local alignment of \( z \) with \( D \) mirrors plausible neural learning mechanisms.
\end{enumerate}

\subsubsection{Summary}
When \( D \) is the sigmoid function, \( H_{\text{SKA}} \) matches \( H_{\text{Shannon}} \) exactly, with a difference of zero. This result, derived from the continuous entropy \( H_{\text{SKA}} = -\frac{1}{\ln 2} \int z \, dD \), validates SKA’s foundation and its seamless integration with information theory, leveraging continuous dynamics for neural learning with classical information theory.
	

\subsection{The Fundamental Law of Entropy Reduction}

The SKA framework establishes a fundamental law governing how entropy decreases as structured knowledge evolves. This section derives this law using continuous dynamics, reflecting the continuous nature of decision probabilities and knowledge accumulation introduced in Sections 2 and 3. We then provide a discrete approximation for practical implementation, ensuring the framework’s applicability to neural networks while rooting it in a continuous theoretical foundation.

\subsubsection{Continuous Dynamics}

For a single neuron, the rate of entropy change with respect to structured knowledge \( z \) is derived from the continuous entropy \( H = -\frac{1}{\ln 2} \int z \, dD \). Taking the partial derivative:
\begin{equation}
	\frac{\partial H}{\partial z} = -\frac{1}{\ln 2} z D (1 - D).
\end{equation}
This follows from \( D = \frac{1}{1 + e^{-z}} \) (as derived in Section 4), where \( \frac{dD}{dz} = D (1 - D) \), and reflects the neuron’s local contribution to entropy reduction. For a layer \( l \) with \( n_l \) neurons at step \( k \), this extends to each neuron \( i \):
\begin{equation}
	\frac{\partial H^{(l)}}{\partial z_i^{(l)}(k)} = -\frac{1}{\ln 2} z_i^{(l)}(k) D_i^{(l)}(k) \left(1 - D_i^{(l)}(k)\right).
\end{equation}
Equation (21) governs the continuous reduction of layer-wise entropy \( H^{(l)} \), driven by the alignment of \( z_i^{(l)}(k) \) with the sigmoidal decision probability \( D_i^{(l)}(k) \). This dynamic, localized process underpins the SKA’s forward-only learning mechanism, leveraging the continuous evolution of \( D \) over time or input processing.

\subsubsection{Discrete Dynamics}

In practice, neural networks operate over discrete forward steps. For a single neuron at step \( k \), the entropy gradient approximates the continuous form, incorporating the change in decision probability \( \Delta D_k = D_k - D_{k-1} \):
\begin{equation}
	\frac{\partial H}{\partial z} \big|_k = -\frac{1}{\ln 2} \left[ z_k D_k (1 - D_k) + \Delta D_k \right].
\end{equation}
For layer \( l \) at step \( k \), this becomes:
\begin{equation}
	\frac{\partial H^{(l)}}{\partial z_i^{(l)}(k)} = -\frac{1}{\ln 2} z_i^{(l)}(k) \left[ D_i^{(l)}(k) \left(1 - D_i^{(l)}(k)\right) + \Delta D_i^{(l)}(k) \right].
\end{equation}
Equation (23) adapts continuous law to discrete steps where \( \Delta D_i^{(l)}(k) \) denotes the change in \( D_i^{(l)}(k) \). While (21) captures the ideal case of continuous dynamics, and Equation (23) provides a computable approximation, aligning knowledge adjustments with observed changes in decision probabilities across discrete iterations.
\subsection{Generalization to Fully Connected Networks}

The SKA framework builds from single neurons to fully connected neural networks without losing continuity, utilizing earlier defined principles of continuous entropy reduction. In a \( L \)-layer network, knowledge and decision probabilities develop in a stratified manner, decreasing the overall entropy through local, one-way movements. In this part, I describe the workings of SKA at different levels while keeping the network's biology-inspired scalability in mind.

For a neural network having \(L\) layers:  

\begin{itemize}  

	\item \(\mathbf{D}^{(l)}_k = \sigma(\mathbf{z}^{(l)}_k)\), the decision probabilities derived via the sigmoid function,  

	\item \(z_k^{(l)}\) the knowledge vector at layer \(l\) and step \(k\) is defined as,  

	\item \(\mathbf{D}^{l}\) is defined as decision probabilities of l-th layer of neural network,  
\end{itemize}  

Note: In this form, the neural network is used for static systems where information does not change with time.

The continuous formulation provides the basis for layer-wise entropy which is defined discretely as: 

\begin{equation}
	H^{(l)} = -\frac{1}{\ln 2} \sum_{k=1}^{K} \mathbf{z}^{(l)}_k \cdot \Delta \mathbf{D}^{(l)}_k.
\end{equation}

The coherence relation describing the knowledge and decision shift as measured is given by: 

\begin{equation}
	\mathbf{z}^{(l)}_k \cdot \Delta \mathbf{D}^{(l)}_k = \|\mathbf{z}^{(l)}_k\| \| \Delta \mathbf{D}^{(l)}_k\| \cos(\theta^{(l)}_k).
\end{equation}

Total network entropy across layers is defined as: 

\begin{equation}
	H = \sum_{l=1}^{L} H^{(l)}.
\end{equation}

Progress is made by adjusting \( \mathbf{z}^{(l)}_k \) to \( \Delta \mathbf{D}^{(l)}_k \) at each layer, enabling local reductions of \( H^{(l)} \) without necessitating backpropagation of errors. In the limit of continuity, this smoothing aligns with evolution of knowledge’s alignment, in this case represented with gaps for ease of computation.

\subsection{Learning Without Backpropagation}

The SKA model achieves learning through localized entropy minimization, allowing no backpropagation in favor of forward-only dynamics. This subsection breaks down the weight update procedure and the relevant metrics framed within continuously reducing entropy, with adjustments made for discretized entropy minimization in fully-connected networks.  

The driving force behind the entropy minimization at layer \( l \) is:

\begin{equation}
	\frac{\partial H^{(l)}}{\partial w_{ij}^{(l)}} = -\frac{1}{\ln 2} \sum_{k=1}^{K} \frac{\partial (\mathbf{z}^{(l)}_k \cdot \Delta \mathbf{D}^{(l)}_k)}{\partial w_{ij}^{(l)}}.
\end{equation}
The update rule adjusts weights forward:
\begin{equation}
	w_{ij}^{(l)} \leftarrow w_{ij}^{(l)} - \eta \frac{\partial H^{(l)}}{\partial w_{ij}^{(l)}}.
\end{equation}
Here, \( \Delta D_i^{(l)}(k) \) is computed directly from forward passes, bypassing the need for error backpropagation. This local adjustment aligns with the continuous dynamics of knowledge evolution, approximated over discrete steps.

\subsubsection*{Step-wise Entropy Change}
To quantify knowledge accumulation, the step-wise entropy change at layer \( l \) and step \( k \) is:
\begin{equation}
	\Delta H^{(l)}_k = H^{(l)}_k - H^{(l)}_{k-1} = -\frac{1}{\ln 2} \mathbf{z}^{(l)}_k \cdot \Delta \mathbf{D}^{(l)}_k.
\end{equation}
This measures uncertainty reduction as \( \mathbf{z}^{(l)}_k \) aligns with \( \Delta \mathbf{D}^{(l)}_k \), with total layer entropy as:
\begin{equation}
	H^{(l)} = \sum_{k=1}^{K} \Delta H^{(l)}_k.
\end{equation}

\subsubsection*{Entropy Gradient}
The gradient of \( H^{(l)} \) with respect to \( \mathbf{z}^{(l)}_k \) at step \( k \) is:
\begin{equation}
	\nabla H^{(l)} = \frac{\partial H^{(l)}}{\partial \mathbf{z}^{(l)}_k} = -\frac{1}{\ln 2} \mathbf{z}^{(l)}_k \odot \mathbf{D}'^{(l)}_k - \Delta \mathbf{D}_k^{(l)},
\end{equation}
where \( \mathbf{D}'^{(l)}_k = \mathbf{D}^{(l)}_k \odot (\mathbf{1} - \mathbf{D}^{(l)}_k) \) is the sigmoid derivative. This gradient guides updates to minimize \( H^{(l)} \), aligning knowledge with decision shifts.

\subsubsection*{Knowledge Evolution Across Layers}
The gradient \( \nabla H^{(l)} = -\frac{1}{\ln 2} \mathbf{z}^{(l)}_k \odot \mathbf{D}'^{(l)}_k - \Delta \mathbf{D}_k^{(l)} \) captures the change in entropy for a given layer \( l \) during the \( k^{th} \) iteration. \( \mathbf{D}^{(l-1)}_k \) feeds to \( \mathbf{z}^{(l)}_k \) so each layer can flexibly self-adjust – so the model interprets wide features first, then narrows down on details to refine decision making. This self-driven adjustment resembles a flow of knowledge that has been sampled at discrete intervals.

\subsubsection*{Governing Equation of SKA}
The network evolves according to:
\begin{equation}
	\nabla H^{(l)} + \frac{1}{\ln 2} \mathbf{z}^{(l)}_k \odot \mathbf{D}'^{(l)}_k + \Delta \mathbf{D}_k^{(l)} = 0,
\end{equation}
where \( \nabla H^{(l)} \) reduces entropy for each layer, with updates done by \( -\nabla H^{(l)} \) to steer \( \mathbf{z}^{(l)}_k \) towards \( \Delta \mathbf{D}^{(l)}_k \).

\subsubsection*{Inter-Layer Entropy Change}
The entropy change between layers \( l \) and \( l+1 \) at step \( k \) is:
\begin{equation}
	\Delta H^{(l,l+1)}_k = -\frac{1}{\ln 2} \left[ \mathbf{z}^{(l+1)}_k \cdot \Delta \mathbf{D}^{(l+1)}_k - \mathbf{z}^{(l)}_k \cdot \Delta \mathbf{D}^{(l)}_k \right].
\end{equation}
This describes the movement of knowledge over space and complements the temporal flow provided by \( \nabla H^{(l)} \), while entropy is flowing down the network.
\section{Applications of the Obtained Results}
\subsection{Application to Neural Networks}

SKA organizes information into different layers, weakening total entropy \( H \) through flow-like processes that can be simulated in steps. A multilayer perceptron (MLP) can be trained to minimize \( H \), where \( \cos (\theta^{(l)}_k) \) stands as a suitable indicator for the level of alignment between \( \mathbf{z}^{(l)}_k \) and \(\Delta \mathbf{D}^{(l)}_k \). This method takes advantage of the system's scalability and self-management, which can be utilized in various networks, hence using a forward-only approach.

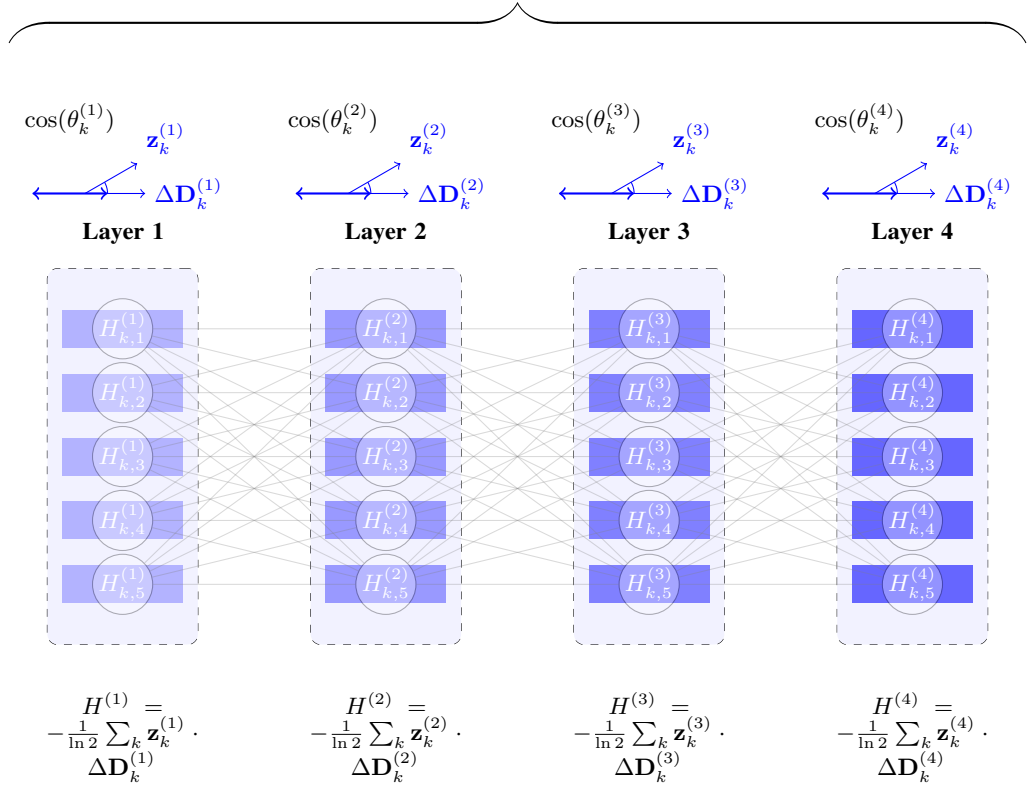
\begin{figure}[ht!]
	\centering
 \begin{tikzpicture}[
		neuron/.style={circle, draw, fill=white, minimum size=0.8cm, opacity=0.3},
		layer/.style={rectangle, draw, dashed, minimum width=2cm, minimum height=5cm, rounded corners},
		entropy/.style={fill=blue!5, rounded corners, minimum width=2cm, minimum height=5cm},
		arrow/.style={-Stealth, thick},
		textnode/.style={align=center, text width=2.4cm, font=\footnotesize}
		]
		\foreach \l/\x in {1/0, 2/3.5, 3/7, 4/10.5} {
			\node[layer] (L\l) at (\x, 0) {};
			\node[entropy] (E\l) at (\x, 0) {};
			\node[above=0.2cm of L\l, textnode] {\textbf{Layer \l}};
			\foreach \i/\y in {1/1.7, 2/0.85, 3/0, 4/-0.85, 5/-1.7} {
				\fill[blue!\fpeval{30+(\l-1)*10}] (\x-0.8, \y-0.25) rectangle (\x+0.8, \y+0.25);
				\node[white, font=\footnotesize\bfseries] at (\x, \y) {$H^{(\l)}_{k,\i}$};
			}
			\foreach \n/\y in {1/1.7, 2/0.85, 3/0, 4/-0.85, 5/-1.7} {
				\node[neuron] (N\l\n) at (\x, \y) {};
			}
			\node[below=0.5cm of L\l, textnode] {
				$H^{(\l)} = -\frac{1}{\ln 2} \sum_{k} \mathbf{z}^{(\l)}_k \cdot \Delta \mathbf{D}^{(\l)}_k$
			};
		}
		\foreach \l [remember=\l as \lastl (initially 1)] in {2,3,4} {
			\foreach \i in {1,...,5} {
				\foreach \j in {1,...,5} {
					\draw[gray, opacity=0.3] (N\lastl\i) -- (N\l\j);
				}
			}
		}
		\foreach \l/\x in {1/1.5, 2/5, 3/8.5, 4/12} { 
			\draw[<->, blue, thick] (\x-2.7, 3.5) -- (\x-1.7, 3.5);
			\node at (\x-2.2, 4.5) [font=\footnotesize] {$\cos(\theta^{(\l)}_k)$};
			\draw[blue] (\x-2, 3.5) ++(0:0.3cm) arc (0:30:0.3cm);
			\draw[->, blue] (\x-2, 3.5) -- ++(0:0.8cm) node[right, font=\footnotesize] {$\Delta \mathbf{D}^{(\l)}_k$};
			\draw[->, blue] (\x-2, 3.5) -- ++(30:0.8cm) node[above right, font=\footnotesize] {$\mathbf{z}^{(\l)}_k$};
		}
		\node[above=5cm of L1.north, align=center] {\large \textbf{Layer-wise Entropy Reduction in SKA}};
		\draw[decorate, decoration={calligraphic brace, amplitude=15pt}, thick]
		($(L1.north west)+(-0.5,3)$) -- ($(L4.north east)+(0.5,3)$)
		node[midway, above=1cm, font=\footnotesize] {$H = \sum_{l=1}^{L} H^{(l)} \downarrow$ as knowledge structures};
	\end{tikzpicture}
	\caption{Layer entropy reduction for a given step \( k \). The entropy level is represented by color intensity, darker blue corresponds to lower entropy. For illustration purpose only, entropy progressively decreases from Layer 1 to Layer 4. Each layer lowers entropy on a local level by coordinating knowledge vector \( \mathbf{z}^{(l)}_k \) with decision change vector \( \Delta \mathbf{D}^{(l)}_k \) as measured by \( \cos(\theta^{(l)}_k) \).}
	\label{fig:network}
\end{figure}
\FloatBarrier
\subsection{Tensor-Based Implementation of SKA}

In order to improve computing performance and scalability, \textbf{tensor-based implementation} is introduced to the SKA framework. This approach preserves the theory of SKA and uses tensor operations to allow for efficient and parallelizable learning. With a tensor approach to SKA's mathematical formulation, we optimize \textbf{knowledge accumulation, entropy reduction, and decision updates} computation on several layers and step $K$.

\subsubsection*{Tensor Definitions}
The operation of a neural network under SKA can be expressed using the following tensors: 
The following tensors can be used to represent a neural network functioning under SKA:

\begin{itemize}
	\item \textbf{Knowledge Tensor} ($\mathbf{Z}$): Represents structured information within each neuron at different \textbf{layers} ($L$), \textbf{steps} ($K$) and \textbf{number of neurons} ($n_{\text{max}}$) -dimension.

	\item \textbf{Decision Probability Tensor} ($\mathbf{D}$): Keeps neuron activations as knowledge values transformed using sigmoid function.
	\item \textbf{Shift Tensor \(\Delta \mathbf{D}\):} Represents \textbf{local probability shifts} of change between steps, aligning with SKA’s entropy-based learning mechanism.

	\item \textbf{Weight Tensor} ($\mathbf{W}$) and \textbf{Bias Tensor} ($\mathbf{b}$): Set knowledge parameters that change over time for enhancing knowledge retention.
\end{itemize}
\subsubsection*{Forward-Only Learning and Knowledge Update}
SKA revises knowledge using only the forward path as shown in below SKA update formula. 

\[
\mathbf{Z} = \mathbf{W} \cdot \mathbf{X} + \mathbf{b}
\]
\noindent \textbf{\textit{$\mathbf{X}$}} refers to the input tensor. As opposed to backpropagation, SKA comes with a notable advantage; SKA does not require a back propagation of gradients, which means less computational resources are consumed.

\subsubsection*{Entropy Computation and Learning}
SKA’s entropy formulation can be naturally expressed using tensor operations:

\[
\mathbf{H} = -\frac{1}{\ln 2} \sum_{k=1}^{K-1} \mathbf{Z}_{k+1} \cdot \Delta \mathbf{D}_k,
\]

where \textbf{entropy decreases layer by layer} as structured knowledge accumulates.

To optimize learning, entropy gradients are computed as:

\[
\nabla \mathbf{H} = -\frac{1}{\ln 2} \mathbf{Z} \odot \mathbf{D}' -\Delta \mathbf{D},
\]

where \textbf{$\mathbf{D}'$} is the derivative of the sigmoid function.

\subsubsection*{Weight Updates and Learning Stability}
Weight changes occur according to an entropy minimization rule:

\[
\mathbf{W} \leftarrow \mathbf{W} - \eta \frac{\partial \mathbf{H}}{\partial \mathbf{W}}.
\]
 
An \textbf{alignment tensor} determines if the knowledge updates deviate from their expected paths:

\[
\Theta_{l,k} = \cos(\theta^{(l)}_{k+1}) = \frac{\mathbf{Z}_{l,k+1} \cdot \Delta \mathbf{D}_{l,k}}{\|\mathbf{Z}_{l,k+1}\| \|\Delta \mathbf{D}_{l,k}\| }.
\]

This alignment mechanism ensures that SKA remains \textbf{structured} and \textbf{controlled}.

\subsubsection*{Advantages of Incorporating Tensors in Mathematical Modeling}

\begin{itemize}

    \item \textbf{Simultaneous Execution of Tasks}: The structure of tensors allows for parallel processing within a single layer or across multiple layers and neurons, optimizing resource usage.
    
    \item \textbf{Flexible Connectivity Patterns}: The arrangement allows designs with additional hidden layers to be included without major alterations to the design process.
    
    \item \textbf{Analogous Low Level Operation}: Structures which only compute the forward pass are compatible with design constraints for power-oriented low-complexity real-time operations.

\end{itemize}

This tensor-based model solidifies SKA’s claim as a \textbf{computationally efficient, biologically plausible, and scalable learning paradigm}. Further work will assess its performance when compared to model versions utilizing traditional backpropagation, investigating applications in edge computing, real-time AI, and neural network optimizations.

\subsection{Entropy Evolution in SKA}

One of the most prominent observations within the SKA framework is the distinctive movement of entropy across layers during learning processes. Unlike standard deep networks with non-uniformly varying dynamics of entropy across layers, SKA shows an extraordinary phenomenon: \textbf{all layers are observed tending towards a singular entropy equilibrium value, converging at a predefined entropy level}. This suggests that entropy is not simply decreasing but organizing itself in some orderly manner.

\subsubsection{Empirical Observation: Layer-Wise Entropy Convergence}

\begin{figure}[ht!]
	\centering
	\includegraphics[width=0.8\textwidth]{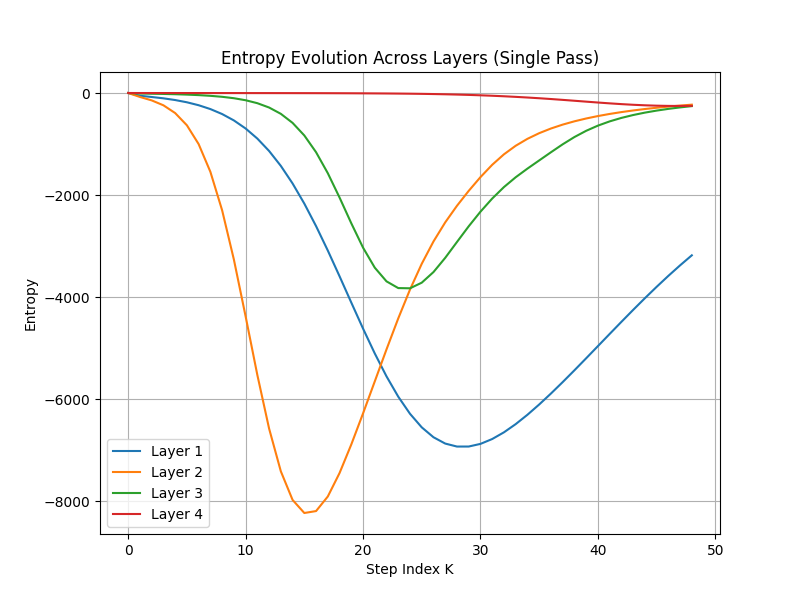}
	\caption{Entropy transformation over layers in an SKA neural network. Layers 2, 3, and 4 achieve a common entropy equilibrium \( K = 49\). Layer 1 converges to this value, albeit at a more gradual rate.}
	\label{fig:entropy_evolution}
\end{figure}

The various layers’ entropy evolution over several forward learning steps is showcased in Figure~\ref{fig:entropy_evolution}. The main highlights are:

\begin{itemize}

	\item \textbf{Layer-Specific Minima}: Local minima for every layer is associated with local minimum entropy value at particular forward steps (\(K\)) after which the value increases gently followed by stabilization.

	\item \textbf{Convergence Toward a Common Equilibrium}: The entropy for layers 2, 3, and 4 approaches a constant value approximately at step \(K=49\) which appears to play a critical equilibrium role.

	\item \textbf{Slow Convergence of Layer 1}: This layer seems to behave similarly when it comes to reducing entropy however, clearly more gradually than the other layers suggesting that they are positioned deeper in the hierarchy of knowledge structuring.

\end{itemize}

All things taken into consideration, structured entropy SKA systems are able to achieve indicate that systems have more to SKA than simply minimizing the total levels of entropy, but rather showing a clear attempt to redistribute it across every layer efficiently.

\FloatBarrier
\subsubsection{Cosine Alignment Evolution}

\begin{figure}[ht!]
	\centering
	\includegraphics[width=0.8\textwidth]{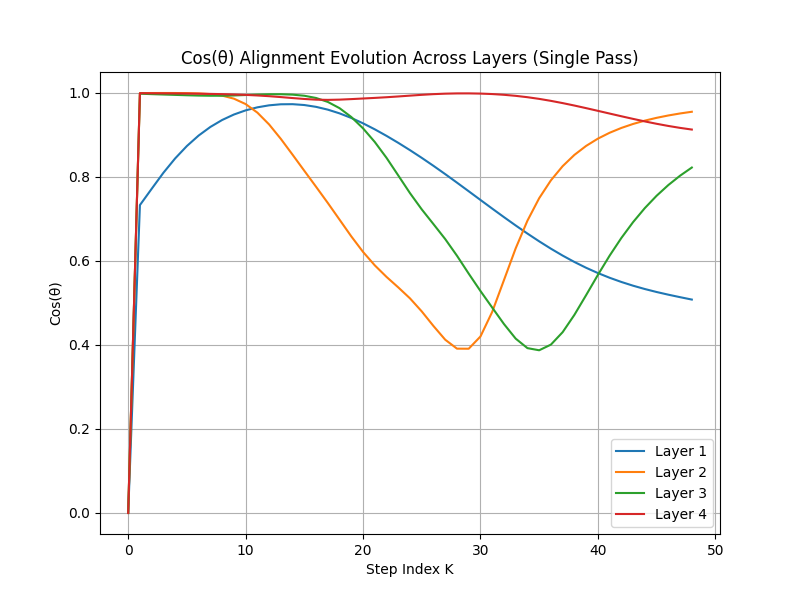}
	\caption{Cosine alignment evolution in SKA. The stabilization of the \(\cos(\theta_k^{(l)})\) over steps indicates organized knowledge alignment across the layers.}
	\label{fig:cosine_evolution}
\end{figure}

Besides studying entropy evolution, SKA also exhibits a remarkable phenomenon known as \textbf{cosine alignment evolution}. As stated before, Figure~\ref{fig:cosine_evolution} depicts how the alignment between knowledge vectors \(\mathbf{z}^{(l)}_k\) and decision probability shift \(\Delta \mathbf{D}^{(l)}_k\) grows over time.  

\begin{itemize}  

	\item \textbf{Cosine Alignment Convergence}: Across layers, \( \cos(\theta^{(l)}_k) \) converges toward a stable value which signals a growing alignment between knowledge accumulation and decision making updates.

	\item \textbf{Parallel Behavior to Entropy}: The exporting of cosine value consistency occurs simultaneously with entropy equilibrium, further confirming the coherent learning strategies of SKA.

	\item \textbf{Layer-Wise Synchronization}: Alignment convergence across layers hints towards the existence of self-organization phenomenon which would optimize knowledge updates via entropy reduction.

\end{itemize}  

This provides further evidence that SKA exhibits a learning behavior independent of
traditional gradient approaches within learning systems that evolve in a structured, sequential manner.
\FloatBarrier
The organized form of the evolution of entropy and cosine alignment offers a new framework concerning learning in neural networks which might be even more useful when attempting to explain how biological and artificial learning systems self-structure their internal representations.
\subsubsection{Output Decision Probability Evolution}

\begin{figure}[ht!]
	\centering
	\includegraphics[width=0.8\textwidth]{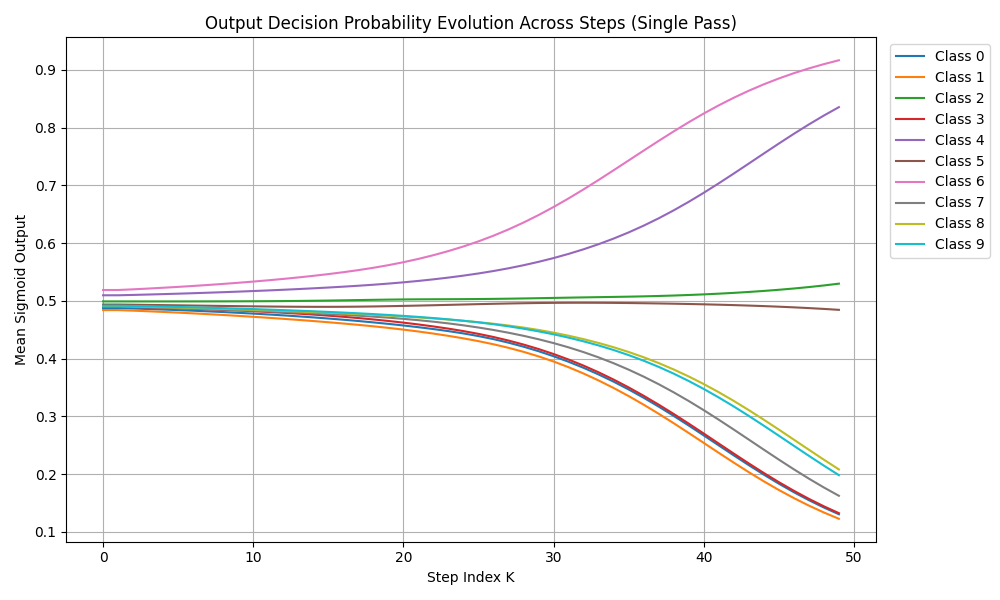}
	\caption{Change in output decision probabilities over forward steps in SKA. Distinctively from other models, SKA improves class distinctiveness incrementally without disrupting defined probability structures.}
	\label{fig:decision_probability}
\end{figure}

Alongside entropy and cosine alignment, one vital visualization in SKA is the \textbf{evolution of output decision probabilities} through forward steps. In Figure~\ref{fig:decision_probability}, we show how the mean decision probability of all 10 classes is distributed as the learning proceeds.

\begin{itemize}

	\item \textbf{Gradual Separation of Classes}: The decision probabilities SKA achieves at the end of the learning period show that the system improves class distinguishability, which is indicative of class separability refinement, even in the absence of explicit gradient updates.  

	\item \textbf{Emergent Stability}: The increase in the number of steps results in a stabilization of the decision probabilities, which signals the ability to reliably reach a systematized form of classification.

	\item \textbf{No Drastic Separation}: In contrast to classical models, SKA does not create sharp decision boundaries. SKA gradually refines knowledge build-up which incorporates soft boundaries in probabilistic terms.</p>

\end{itemize}

This provides additional proof for SKA's self-organizing, entropy-driven classification process, emphasizing the main difference from learning that relies on backpropagation.

\FloatBarrier

\subsubsection{Frobenius Norm Evolution of the Knowledge Tensor}

Apart from tracking entropy, cosine alignment, and output decision probabilities, an informative perspective in SKA is the \textbf{Frobenius norm} of the knowledge tensor \(\mathbf{z}^{(l)}\). For a layer \(l\), the definition of the Frobenius norm is:

\[
\|\mathbf{z}^{(l)}\|_F \;=\; \sqrt{\sum_{i,j} \Bigl(z_{ij}^{(l)}\Bigr)^2},
\]
where \(z_{ij}^{(l)}\) is the knowledge value of neuron \(j\) in sample \(i\) (before the sigmoid activation).

\begin{itemize}
	\item \textbf{Layer-Specific Magnitude Growth}:  
	Each layer’s Frobenius norm reflects the overall magnitude of its knowledge tensor \(\mathbf{z}^{(l)}\). A larger norm indicates that the pre-sigmoid activations are more extreme, suggesting stronger or more polarized responses. Interestingly, while some layers may exhibit rapid increases in their norms, our observations show that the final layer (Layer 4) tends to grow more slowly. This gradual increase in Layer 4’s Frobenius norm suggests that, despite its role in driving the output logits, its activations remain relatively moderate—possibly indicating an early stabilization of the output during the SKA learning process.
	
	\item \textbf{Single-Pass Dynamics}:  
	Under a single-pass, forward-only scheme, some layers may exhibit steadily increasing norms, reflecting the absence of a backward error signal that would typically constrain large activations. This effect can be particularly pronounced in the final layer, where classification logits may grow larger as the model strives to minimize local entropy.
	
	\item \textbf{Relationship to Entropy and Alignment}:  
	While entropy and cosine alignment measure how well the knowledge tensor \(\mathbf{z}^{(l)}\) aligns with the decision shifts \(\Delta \mathbf{D}^{(l)}\), the Frobenius norm focuses solely on the \emph{magnitude} of \(\mathbf{z}^{(l)}\). Thus, a layer may have a large norm yet still maintain low entropy if its knowledge vectors are well-aligned with the decision shifts.
\end{itemize}

\begin{figure}[ht!]
	\centering
	\includegraphics[width=0.8\textwidth]{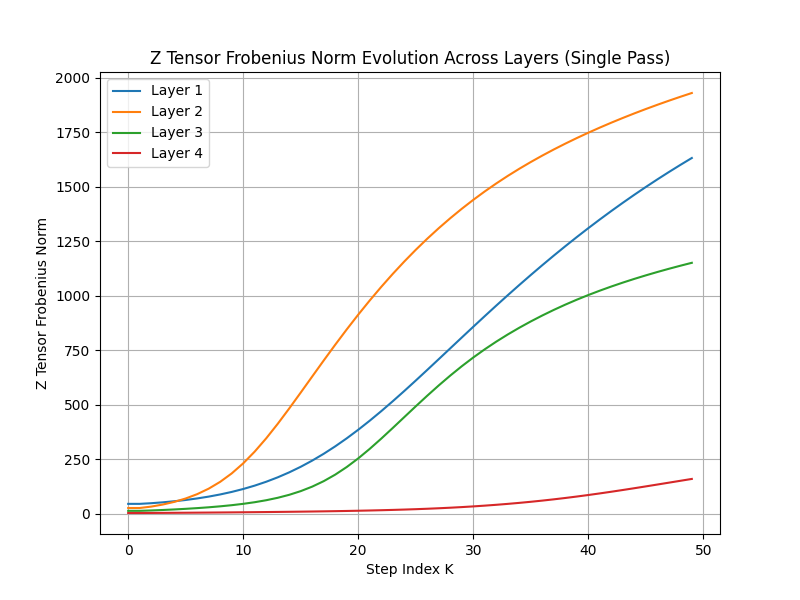}
	\caption{Frobenius norm evolution of the knowledge tensors \(\mathbf{z}^{(l)}\) throughout the layers during single-pass SKA training. It is often the case that the layer with the greater norm will have stronger activations because the updates are only being done in a forward manner and are more focused on local entropy minimization.}
	\label{fig:frobenius_norm_evolution}
\end{figure}

As shown in Figure~\ref{fig:frobenius_norm_evolution}, the knowledge tensors’ Frobenius norms change across several forward steps. It’s particularly interesting that layers may grow or converge at different rates, which provides information on how much each layer is “pushing” its logits to minimize local uncertainty.

\FloatBarrier

In general, monitoring the Frobenius norm offers a new perspective on \textbf{how much magnitude} volumetric factorization captures knowledge at each layer. Combined with entropy, cosine alignment, and evolution of decision probabilities, it adds yet another dimension to how these metrics portray the self-organization of internal representations within SKA networks.

\subsubsection{Entropy Trajectories}

\begin{figure}[ht!]
	\centering
	\includegraphics[width=0.8\textwidth]{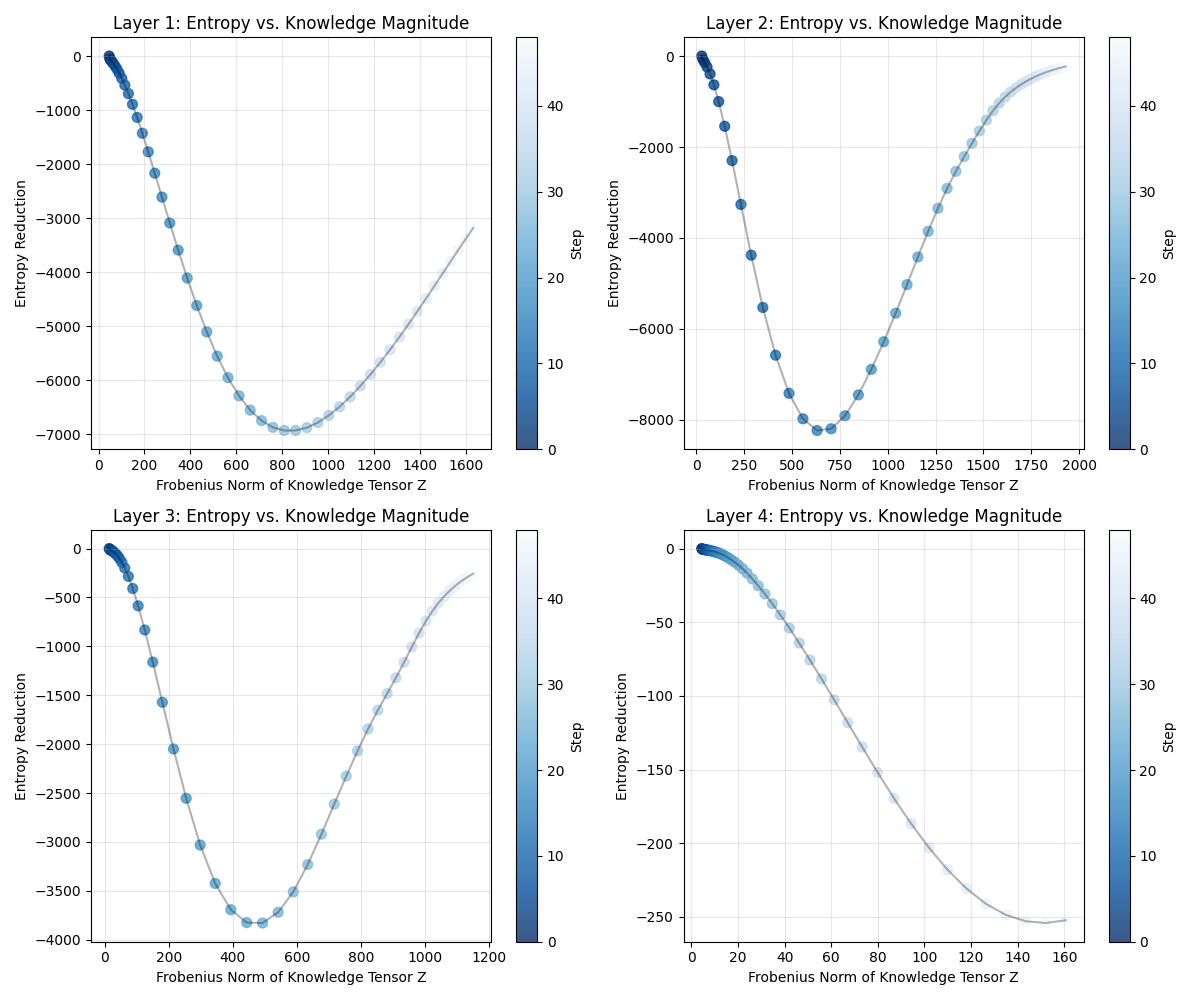}
	\caption{Entropy trajectories over the layers, tracked with respect to the Frobenius norm of the knowledge tensor. Each layer displays a U-shaped curve of entropy, with the minimum shifting towards lower values of knowledge magnitude as depth increases.}
	\label{fig:entropy_vs_frobenius}
\end{figure}

An important observation in SKA is the existence of organized entropy trajectories, especially when graphed against knowledge magnitude. Figure~\ref{fig:entropy_vs_frobenius} demonstrates how reduction of entropy relates to the Frobenius norm of the knowledge tensor at different layers.  

\begin{itemize}  

\item \textbf{U-Shaped Relationship}: Every layer displays a typical U-shaped pattern, with entropy reduction exhibiting decrease with respect to knowledge, followed by an increase after reaching a minimum.  

\item \textbf{Progressive Shift of the Minimum}: The entropy minimum occurs at a progressively lower knowledge magnitude as we move from Layer 1 to Layer 4. This indicates that the lower layers are less efficient in utilizing knowledge magnitude to achieve entropy minimization.  

\item \textbf{Monotonic Knowledge Growth}: Unlike most models that impose active bounds or regularization on knowledge,
SKA allows these mechanisms to arise intrinsically without any tuning needed.  

\end{itemize}  

This kind of behavior suggests that entropy minimization is an outcome of control over the dynamics of knowledge accumulation, which strengthens the claim of SKA being self-organizing system.

\FloatBarrier

The finding demonstrates a new facet on how entropy-driven learning unfolds in SKA: each layer organizes knowledge on its own, without any external molding.

\subsubsection{Theoretical Interpretation}

The equilibrium of the entropy as we observed it is consistent with the SKA principle of knowledge alignment drives learning SKA  inference. The structure of learning is self-organized in the following sense: the layers of knowledge are driven towards equilibrium, while accumulation of knowledge is balanced. 

This indicates the possible existence of a fundamental law governing SKA-based neural networks.

\begin{quote}
	\textit{In an SKA neural network, layer-wise entropy converges to an equilibrium state where knowledge accumulation stabilizes across hierarchical representations.}
\end{quote}

\section{Conclusion and Future Works}

The SKA framework redefines neural learning as a process of entropy-guided knowledge organization. This approach presents a foundational shift from conventional gradient-based training. By formulating entropy as a continuous, dynamic accumulation of knowledge, given as

\begin{equation}
	H = -\frac{1}{\ln 2} \int z \, dD.
\end{equation}

We have demonstrated that the sigmoid activation function emerges naturally from entropy minimization principles which provide a biologically plausible and mathematically elegant basis for learning without backpropagation.

The SKA framework enables layer-wise optimization, where each layer independently aligns knowledge vectors with decision probability shifts. This local learning dynamic not only decentralizes the training process but also enhances interpretability through angular alignment metrics such as \( \cos(\theta^{(l)}_k) \). Entropy progressively decreases across layers, compressing knowledge representations while maintaining information fidelity, and offering a scalable architecture suitable for real-time and resource-constrained applications.

As research on the SKA framework progresses, it holds the potential to transform the landscape of artificial intelligence by aligning learning mechanisms more closely with natural information systems. The interplay between entropy, structure formation, and local decision dynamics hints at a deeper organizational principle underlying both artificial and biological learning. In this light, SKA becomes not merely a training methodology but a paradigm through which the self-organization of intelligent behavior can be understood and replicated.

Looking ahead, future research will focus on extending the SKA framework to real-time domains such as visual and auditory processing, where continuous adaptation and forward-only computation offer tangible advantages. Comparative studies with traditional gradient-based methods on benchmark datasets will further elucidate its performance and generalization capabilities. The inherent interpretability of entropy-guided alignment also opens promising directions for transparent AI, where knowledge flow within networks can be tracked and understood. Furthermore, exploring its applicability in distributed systems, neuroscience-inspired architectures, and unsupervised learning could unlock new dimensions in scalable and interpretable machine intelligence.

\section*{Data Availability}
The data that support the findings of this study are available from the corresponding author upon reasonable request.
\section*{Competing Interests}
The authors declare that there are no competing interests.

\section*{Author Contributions}
The author confirms sole responsibility for all aspects of this work, including conceptualization, methodology, analysis, writing, and manuscript preparation.

\section*{Ethics Declaration}
Not applicable.
\bibliographystyle{unsrtnat}


\end{document}